\documentclass[letterpaper]{article}
\usepackage{graphicx}
\usepackage{caption}
\usepackage{algorithm}
\usepackage{algorithmic}
\title{On the Unlikelihood of D-Separation}
\usepackage{authblk}
\author{Itai Feigenbaum}
\author{Huan Wang}
\author{Shelby Heinecke}
\author{Juan Carlos Niebles}
\author{Weiran Yao}
\author{Caiming Xiong}
\author{Devansh Arpit}
\affil{Salesforce AI Research}
\date{}

\usepackage{amsmath}
\usepackage{graphicx}
\usepackage{amsthm}
\usepackage{amsfonts}
\newtheorem{theorem}{Theorem}

\newtheorem{lemma}{Lemma}
\newtheorem{definition}{Definition}
\theoremstyle{definition}

\usepackage{tikz}
\usetikzlibrary{decorations.pathreplacing}
\usetikzlibrary{fadings}
\usepackage{xcolor}
\usepackage{caption}
\usepackage{subcaption}
\usepackage{cancel}

\usepackage[title]{appendix}
\usepackage{thmtools}
\usepackage{thm-restate}

\begin{document}

\maketitle

\begin{abstract}
Causal discovery aims to recover a causal graph from data generated by it; constraint-based methods do so by searching for a d-separating conditioning set of nodes in the graph via an oracle. In this paper, we provide analytic evidence that on large graphs, d-separation is a rare phenomenon, even when guaranteed to exist. Our analysis implies poor average-case performance of existing constraint-based methods, except on a vanishingly small class of extremely sparse graphs. We consider a set $V=\{v_1,\ldots,v_n\}$ of nodes, and generate a random DAG $G=(V,E)$ where $[v_i,v_j] \in E$ with i.i.d. probability $p_1$ if $i<j$ and $0$ if $i > j$. We provide upper bounds on the probability that a subset of $V-\{x,y\}$ d-separates $x$ and $y$, under different subset selection scenarios, conditional on $x$ and $y$ being d-separable; our upper bounds decay exponentially fast to $0$ as $|V| \rightarrow \infty$ for any fixed expected density. We then analyze the average-case performance of constraint-based methods, including the PC Algorithm, a variant of the SGS Algorithm called UniformSGS, and also any constraint-based method limited to small conditioning sets (a limitation which holds in most of existing literature). We show that these algorithms must either suffer from low precision or exponential running time on all but extremely sparse graphs.
\end{abstract}

\section{Introduction}

Causal discovery aims to reverse engineer a {\it causal graph} from the data it generates. The existence of an edge between two nodes can be shown to be equivalent to a property called {\it d-separation} (Definition \ref{def:sep}), defined relatively to two nodes and a subset of the other nodes called a {\it conditioning set}. Constraint-based methods assume access to a d-separation oracle, which deduces d-separation from the data, and discover the graph via a series of oracle calls \cite{glymour2019review}. Perhaps the most well-known constraint-based method is the {\it PC Algorithm}, which is a specialization of the {\it SGS Algorithm}  \cite{spirtes1991algorithm}. Under some assumptions, for sufficiently sparse graphs, the PC Algorithm recovers the undirected skeleton of the causal graph correctly and makes at most a polynomial number of oracle calls. This worst-case guarantee fails for non-sparse graphs, but to the best of our knowledge, there is no rigorous analysis of the average case in literature.

\begin{figure}
\includegraphics[width=\linewidth,trim={0 0 0 40pt},clip]{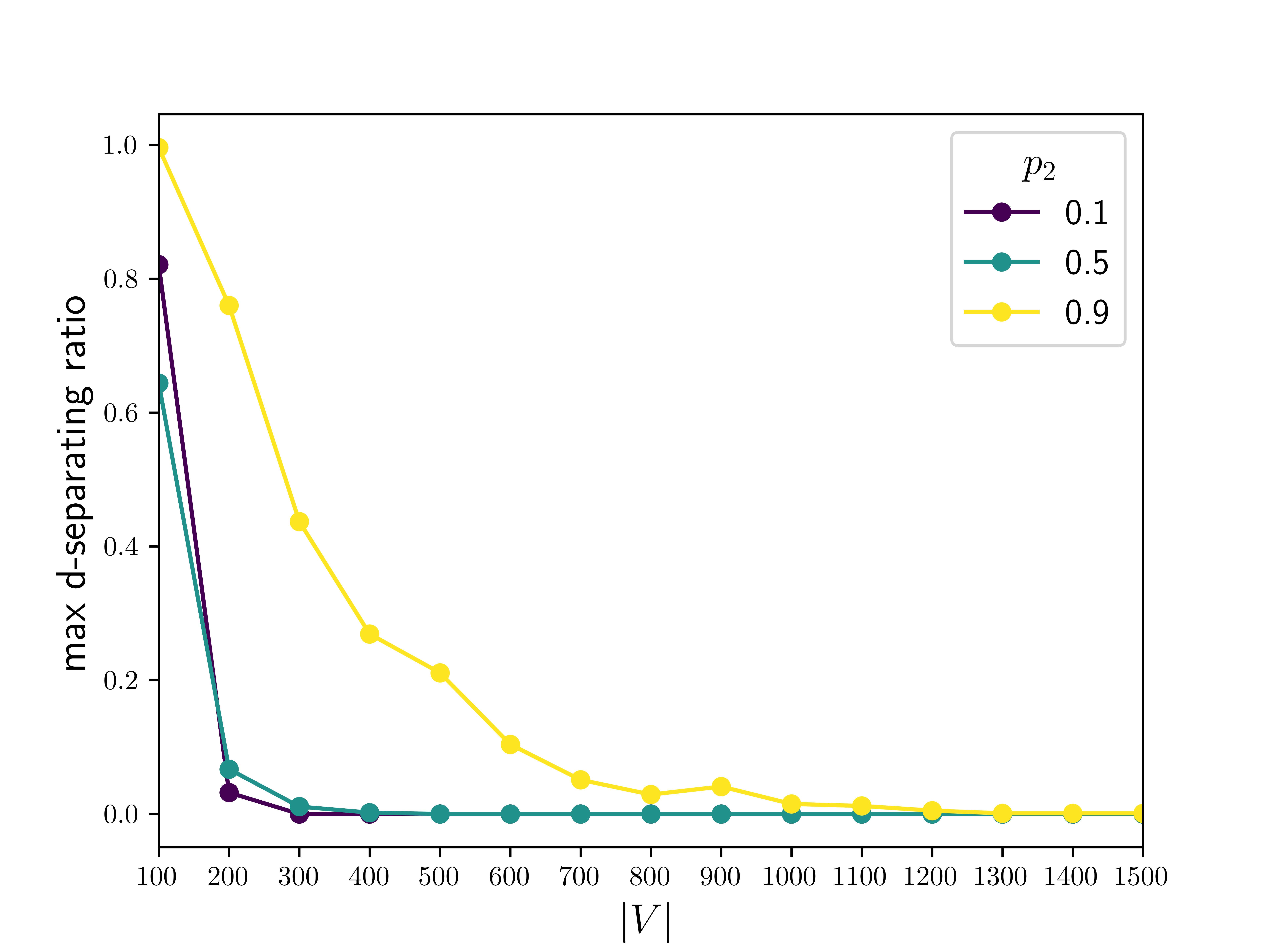}
\caption{d-separation probability. Given a random DAG with $|V|$ nodes and expected density $p_1=0.05$, we sample uniformly at random $100$ d-separable variable pairs. For each pair, we sample $1000$ conditioning sets, including each variable (except the pair) in the conditioning set with i.i.d. probability $p_2$, and then calculate the ratio of the number of d-separating sets over all $1000$ sets. The y-axis is the \underline{maximum} ratio over all $100$ pairs. Consistently with Theorem \ref{thm:randomz}, d-separation quickly becomes rare as $|V| \rightarrow \infty$ despite the low density.} 
\label{fig:data}
\end{figure}

Since searching for d-separation is the core of all constraint-based methods, in this paper we set out to explore the search space. We consider a randomly generated directed acyclic graph (DAG) with $|V|$ nodes and any expected density $p_1$. For any two d-separable nodes, we provide upper bounds on the probability that a conditioning set d-separates the two nodes, in different scenarios. First, we bound the probability of d-separation when each node is included in the conditioning set with any fixed i.i.d. probability $p_2$; second, we bound the probability that there exists a d-separating set of size at most some linear fraction of $|V|$; and third, we bound the probability that a randomly chosen conditioning set of any fixed size is d-separating. All of our bounds are $O(e^{-|V|})$ for any fixed $p_1$. In Figure \ref{fig:data}, we give an empirical sense of the first scenario. We refer the reader to the appendix for additional empirical results which confirm the quick decay implied by our bounds.
The informal conclusion from our bounds is that for large $|V|$, d-separation is a very rare phenomenon even when guaranteed to occur, unless the graph is extremely sparse; this gives some indication of the difficulty in finding d-separation.

After establishing our bounds, we analyze their implications for the average-case performance of constraint-based methods. We begin by showing that the {\it extremely sparse} case where $\lim_{|V| \rightarrow \infty}{p_1}=0$ includes only a vanishingly small portion of all possible graphs. In addition to this theoretical argument, we note that sparsity is considered an unrealistic assumption in some fields such as epidemiology \cite{petersen2022causal} and finance \cite{shirokikh2022networks}. We then provide an average-case analysis of constraint-based methods on non extremely sparse large graphs, showing:
\begin{enumerate}
\item[(1)] Any constraint-based method which is restricted to considering only small conditioning sets (sublinear in $|V|$) suffers from low precision. This restriction holds in most of literature: in practice the d-separation oracle is replaced with a statistical conditional independence test, and such tests usually perform badly when the conditioning set is not small \cite{runge2019detecting,li2020nonparametric}.
\item[(2)] Even without externally imposing a conditioning set size restriction, the PC Algorithm suffers either from low precision or exponential running time. While PC's poor performance on non-sparse graphs has been empirically observed \cite{petersen2022causal}, theoretical justification in literature has been limited to worst-case analysis: our average-case results provide a more comprehensive and convincing theoretical justification for this phenomenon.
\item[(3)] We consider a variant of the SGS Algorithm we call UniformSGS that samples conditioning sets uniformly at random without replacement. We show that even when there exists a d-separating conditioning set, UniformSGS takes an expected exponential time to find one.
\end{enumerate}

The remainder of the paper is organized as follows. In Section \ref{subsec:related}, we discuss related literature. In Section \ref{sec:definitions}, we define our model and the relevant terminology. In Section \ref{sec:Bounds}, we provide three bounds on the probability that a conditioning set d-separates two nodes, in different scenarios. In Section \ref{sec:Algorithms}, we show that the extremely sparse case is rare, and analyze the implications of our bounds for constraint-based methods on non extremely sparse graphs. In Section \ref{sec:conclusion} we summarize our results. {\bf All omitted proofs and empirical results are in the appendix}.

\subsection{Related Work}\label{subsec:related}

The early causal discovery approaches can be broadly categorized into two classes, including constrained-based ones such as PC \cite{spirtes1991algorithm} and FCI \cite{spirtes2013causal}, etc. and scored-based ones such as GES \cite{chickering2002optimal}. It was observed that under faithfulness and causal Markov assumptions \cite{SGS93}, a Markov equivalence class of causal graphs could be recovered by exploiting the conditional independence relationships among the observed variables. The causal Markov condition and the faithfulness assumption ensure that there is a correspondence between d-separation properties \cite{verma1988influence} in the underlying causal structure and statistical independence properties in the data. This led to the development of the constraint-based approach to causal discovery, which produces an equivalence class that may contain multiple DAGs or other graphical objects that encode the same conditional independence relationships. Since then, the field of causal discovery has grown significantly. D-Separation and different variations of the PC (and more generally, SGS) Algorithm \cite{spirtes1989causality, spirtes1991algorithm} are at the heart of constraint-based methods of causal discovery, and are the focus of this paper. 

In contrast, the score-based approach \cite{glymour2019review,spirtes2016causal,heinze2018causal,vowels2022d} searches for the equivalence class that yields the highest score under certain scoring criteria \cite{chickering2002optimal}, such as the Bayesian Information Criterion (BIC), the posterior of the graph given the data \cite{heckerman2006bayesian}, or the generalized score functions \cite{huang2018generalized}. Another set of methods is based on functional causal models (FCMs), which represent the effect as a function of the direct causes together with an independent noise term. The causal direction implied by the FCM is recovered by exploiting the model assumptions, such as the independence between the noise and the cause, which holds only for the true causal direction and is violated for the wrong direction, or minimal change principles \cite{ghassami2018multi,scholkopf2021toward}, which states that with correct causal factorization, only a few factors may change under different contexts. Examples of FCM-based causal discovery algorithms include the linear non-Gaussian acyclic model LiNGAM \cite{shimizu2006linear}, the additive noise model ANM \cite{hoyer2008nonlinear}, and the post-nonlinear causal model PNL \cite{zhang2012identifiability}. Recently, causal discovery under changing causal relations \cite{huang2020causal} or with (many) hidden confounders \cite{xie2020generalized,huang2022latent,yao2022temporally} have also been explored. 

\section{Setup and Definitions}\label{sec:definitions}

We denote a directed edge from node $a$ to node $b$ as $[a,b]$, and the underlying undirected edge with $(a,b)$. By {\it path} we specifically mean an undirected simple path; by {\it length} of a path we mean the number of edges in it. Let $V=\{v_1,\ldots,v_n\}$. Let $G=(V,E)$ be a random DAG generated as follows: if $i<j$, $[v_i,v_j] \in E$ with probability $0<p_1<1$, and if $i \geq j$ $[v_i,v_j] \notin E$ deterministically. The generation of each edge is independent of the others. Let $G^*$ be the set of all possible DAGs with nodes $V$ which respect the topological order $v_1,\ldots,v_n$. Note that $G$ is a discrete random variable with support in $G^*$ (every DAG in $G^*$ is generated with nonzero probability); the special case of $p_1=0.5$ corresponds to $G$ being uniform over $G^*$. For any $v_i, v_j \in V$, define $V_{i,j}=V-\{v_i,v_j\}$.

Next, we define colliders, noncolliders, blocking and d-separation, as well as pseudoblocking and pseudoseparation. Our definitions of the first four are consistent with those of Pearl \cite{pearl2009causality}, while the last two are new simple concepts we are introducing here.
\begin{definition}[Collider/Noncollider Path]\label{def:colpath}
Let $P$ be a length $2$ path: specifically, $P$ is a digraph with three nodes $v_i$, $v_k$, $v_j$ and undirected edges $(v_i,v_k)$, $(v_k,v_j)$, which we also denote with $v_i-v_k-v_j$. $P$ is called a {\it collider path} (Figure \ref{fig:collider}) if the directions of the edges are $[v_i,v_k]$, $[v_j,v_k]$ and a {\it noncollider path} (Figure \ref{fig:noncollider}) otherwise. The node $v_k$ is called the {\it middle node} of the path.
\end{definition}
\begin{figure}
	\centering
	\begin{tikzpicture}[shorten >=1pt,]
		\tikzstyle{unit}=[draw,shape=circle,minimum size=0.2cm]
        \node[unit](i) at (0,0)[scale=1]{$v_i$};
        \node[unit](k) at (3,0)[scale=1]{$v_k$};
        \node[unit](j) at (6,0)[scale=1]{$v_j$};
        \draw[->] (i) -- (k);
        \draw[->] (j) -- (k);
	\end{tikzpicture}
	\caption{A collider.}
	\label{fig:collider}
\end{figure}
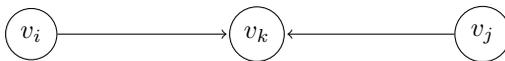
\begin{figure}
        \centering
	\begin{tikzpicture}[shorten >=1pt]
		\tikzstyle{unit}=[draw,shape=circle,minimum size=0.2cm]
        \node[unit](i) at (0,0)[scale=1]{$v_i$};
        \node[unit](k) at (3,0)[scale=1]{$v_k$};
        \node[unit](j) at (6,0)[scale=1]{$v_j$};
        \draw[->] (i) -- (k);
        \draw[->] (k) -- (j);
        \node[unit](i) at (0,2)[scale=1]{$v_i$};
        \node[unit](k) at (3,2)[scale=1]{$v_k$};
        \node[unit](j) at (6,2)[scale=1]{$v_j$};
        \draw[->] (k) -- (i);
        \draw[->] (j) -- (k);
        \node[unit](i) at (0,4)[scale=1]{$v_i$};
        \node[unit](k) at (3,4)[scale=1]{$v_k$};
        \node[unit](j) at (6,4)[scale=1]{$v_j$};
        \draw[->] (k) -- (i);
        \draw[->] (k) -- (j);
	\end{tikzpicture}
	\caption{Noncolliders.}
	\label{fig:noncollider}
\end{figure}
\begin{definition}[Collider/Noncollider Node]\label{def:colnode}
Let $P$ be an arbitrary length path between nodes $a, b \in V$. Every node $c$ in the path s.t. $c \notin \{a,b\}$ is the middle node of a unique length $2$ sub-path in $P$, $P'$, which includes $c$ and the two edges adjacent to $c$ in $P$. We call $c$ a {\it collider (resp. noncollider) node in $P$} iff $P'$ is a collider (resp. noncollider).
\end{definition}
\begin{definition}[Blocking and Pseudoblocking]\label{def:block}
Let $Z \subseteq V$. A path is {\it blocked} by $Z$ iff at least one of the following holds:
\begin{itemize}
    \item The path contains a collider node s.t. the collider and all of the collider's descendants in $G$ are not in $Z$.
    \item The path contains a non-collider node in $Z$.
\end{itemize}
A path is {\it pseudoblocked} by $Z$ iff at least one of the following holds:
\begin{itemize}
    \item The path contains a collider node not in $Z$.
    \item The path contains a non-collider node in $Z$.
\end{itemize}
\end{definition}
\noindent Blocking differs from pseudoblocking only in the requirement that the collider's descendants are not in $Z$. Pseudoblocking is weaker than blocking: every blocked path is pseudoblocked, but not every pseudoblocked path is blocked. Similarly, in the following definition, pseudoseparation is weaker than d-separation. We will only apply the notion of pseudoblocking to length $2$ paths, although it is defined for any length.
\begin{definition}[D-separation and pseudoseparation]\label{def:sep}
Let $v_i, v_j \in V$ s.t. $i \neq j$, and assume $Z \subseteq V_{i,j}$. $v_i$ and $v_j$ are {\it d-separated} by $Z$ iff every path between $v_i$ and $v_j$ is blocked by $Z$. $v_i$ and $v_j$ are {\it pseudoseparated} iff every \underline{length $2$} path between $v_i$ and $v_j$ is \underline{pseudoblocked}. We denote $v_i$ and $v_j$ being d-separated by $Z$ in a graph $G'$ by $v_i \models_{Z,G'} v_j$, and pseudoseparated by $v_i \models_{Z,G'}^{ps} v_j$
\end{definition}
\noindent Note that d-separation requires blocking every path between $v_i$ and $v_j$, and pseudoseparation only requires pseudoblocking the length $2$ paths. Pseudoseparation is useful because length $2$ paths between $v_i$ and $v_j$ are edge disjoint and also middle node disjoint; combined with the fact that pseudoblocking---unlike  blocking---ignores colliders' descendants, it allows for a decoupled analysis of the paths. As we will see, despite pseudoseparation being a weaker requirement than d-separation, it allows us to derive fairly strong bounds. Lemma \ref{lem:weaker} formally establishes the relationship between pseudoseparation and d-separation (the proof is immediate from definitions):
\begin{lemma}\label{lem:weaker}
For any $v_i, v_j \in V$ and $Z \subseteq V_{i,j}$, $v_i \models_{Z,G} v_j \Rightarrow v_i \models_{Z,G}^{ps} v_j$.
\end{lemma}
\noindent The reason d-separation is useful for causal discovery is due to the fact that---under some assumptions---it can be detected via conditional independence testing on the data, as well as the following known result \cite{pearl2009causality}:
\begin{theorem}\label{thm:dsepequiv}
$x, y \in V$ are d-separable in $G$ iff $(x, y) \notin E$.
\end{theorem}
\noindent Throughout this paper, the basic idea is to use Lemma \ref{lem:weaker} to {\bf establish upper bounds on the probability of d-separation by establishing upper bounds on the probability of pseudoseparation}.

\section{Bounds on the Probability of D-Separation}\label{sec:Bounds}

First, some informal intuition: all of our bounds (Theorems \ref{thm:randomz}, \ref{thm:boundedsize} and \ref{thm:fixedsize}) rely on identifying a subset $S$ of length $2$ paths s.t. all paths in $S$ must be pseudoblocked for pseudoseparation (and hence, by Theorem \ref{thm:dsepequiv}, for d-separation) to hold. Furthermore, we choose $S$ so that $|S|$ grows linearly with $|V|$. This is useful because the middle node of each path in $S$ either must be included in or must be excluded of the conditioning set for pseudoseparation to hold. Due to the established linear growth of $|S|$, the number of inclusion/exclusion decision combinations for the middle nodes grows exponentially with $|V|$, however only at most one of those combinations can yield pseudoseparation.

\subsection{Preliminaries}

For this section, let $v_i, v_j \in V$ and assume w.l.o.g. that $i<j$. For every $v_k \in V_{i,j}$, let $P_k$ be the undirected path $v_i-v_k-v_j$, and let $P_{i,j}=\{P_k:v_k \in V_{i,j}\}$ be the set of all potential length $2$ paths between $v_i$ and $v_j$ (regardless of whether they actually exist in the graph). We say that a path $P_k$ {\it exists} in a graph $g=(V,E_g) \in G^*$ if both $(v_i,v_k) \in E_g$ and $(v_j,v_k) \in E_g$; when we discuss existence/inexistence of a path without referring to a particular graph, the underlying graph in question is assumed to be $G$. We prove upper bounds on the probability that a set of nodes $Z$ satisfies $v_i \models_{Z,G} v_j$. More precisely, we prove our upper bounds conditional on $v_i$ and $v_j$ being d-separable in $G$; the implied upper bound on the unconditional probability of d-separation is even tighter. We begin with a few lemmas:
\begin{lemma}\label{lem:independence}
For each $v_k \in V_{i,j}$, let $A_k$ be the event that $P_k$ exists in $G$. Then $\mathbb{P}(A_k)=p_1^2$ for all $v_k \in V_{i,j}$. Furthermore, denoting the event that $(v_i,v_j) \in E$ as $A_{i,j}$, the events $\{A_k:v_k \in V_{i,j}\} \cup \{A_{i,j}\}$ are mutually independent.
\end{lemma}
\begin{proof}
$A_k$ is the event where $(v_i,v_k),(v_k,v_j) \in E$: each of these edges exists with probability $p_1$ and therefore $\mathbb{P}(A_k)=p_1^2$. All paths in $P_{i,j}$ are edge disjoint, and therefore the existence of edges in each of these paths is independent of the existence of edges in the others; in addition, the edge $(v_i,v_j)$ is not in any of the paths in $P_{i,j}$, so its existence is independent of the existence of those paths.
\end{proof}
\begin{lemma}\label{lem:indices}
$P_k$ is a collider path iff $k>j$.
\end{lemma}
\begin{proof}
$P_k$ is a collider iff both edges are directed into $v_k$, which happens iff $k>\max{\{i,j\}}=j$.
\end{proof}
Next, we introduce a few additional useful definitions.
\begin{definition}
Define $V_{nc}=\{v_k \in V_{i,j}:k<j\}$, $V_{c}=\{v_k \in V_{i,j}:k>j\}$, $Q_{nc}=\{P_k:v_k \in V_{nc}\}$ and $Q_c=\{P_k:v_k \in V_c\}$. Note that by Lemma \ref{lem:indices}, $v_k \in V_{nc}$ iff $P_k$ is a noncollider path, and $v_k \in V_c$ iff $P_k$ is a collider path; thus, $Q_{nc}$ and $Q_c$ is the set of all potential noncollider and collider paths respectively. Therefore $|V_{nc}|=|Q_{nc}|=j-2$ (corresponding to $k<j$ and $k \neq i$) and $|V_c|=|Q_c|=|V|-j$ (corresponding to $k>j$).
\end{definition}
\begin{definition}
For $g \in G^*$, denote the number of {\it existing} noncollider paths as $B_{nc}^g$ and the number of existing collider paths as $B_c^g$. When $g=G$, we write $B_{nc}$ and $B_c$ instead of  $B_{nc}^G$ and $B_c^G$.
\end{definition}
\begin{lemma}\label{lem:binom}
$B_{nc}$ is a binomial random variable with parameters $p_1^2$ and $j-2$, while $B_c$ is a binomial random variable with parameters $p_1^2$ and $|V|-j$. Furthermore, $B_{nc}$ and $B_c$ are independent of each other and of the event $(v_i,v_j) \in E$.
\end{lemma}
\begin{proof}
By Lemma \ref{lem:indices}, there are $j-2$ potential noncolliders $Q_{nc}$ and $|V|-j$ potential colliders $Q_c$, and by Lemma \ref{lem:independence} each exists with probability $p_1^2$ independently of the others and of $(v_i,v_j) \in E$.
\end{proof}

\subsection{Bounds}

We turn our attention to establishing the bounds on d-separation probability. Our first bound is for a random conditioning set which includes each node with a fixed i.i.d. probability. $S$ in this proof includes all length $2$ paths between $v_i$ and $v_j$ in $G$.
\begin{theorem}\label{thm:randomz}
Let $Z$ be chosen randomly from $2^{V_{i,j}}$ as follows: for every $v \in V_{i,j}$, we include $v \in Z$ with i.i.d. probability $0<p_2<1$. Then, $\mathbb{P}(v_i \models_{Z,G} v_j|(v_i,v_j) \notin E)$ is upper bounded by $(1-p_1^2+(1-p_2)p_1^2)^{|V|-j}(1-p_1^2+p_2p_1^2)^{j-2}$.
\end{theorem}
\begin{proof}
Let $G^*_{i,j}=\{g=(V,E_g) \in G^*: (v_i,v_j) \notin E_g\}$ be the set of all graphs in which the edge $(v_i,v_j)$ is excluded. Let $g=(V, E_g) \in G^*_{i,j}$ be any particular such graph. Consider some $k$ s.t. the path $P_k$ exists in $g$. $Z$ pseudoblocking $P_k$ is equivalent to $v_k \in Z \Leftrightarrow v_k \in V_{nc}$, that is $v_k$ should be included in $Z$ iff $P_k$ is a noncollider path. We know that $\mathbb{P}(v_k \in Z)=p_2$, and therefore $P_k$ is pseudoblocked with probability $p_2$ if $P_k \in Q_{nc}$, and with probability $(1-p_2)$ if $P_k \in Q_c$. As the inclusion/exclusion decision for each node in/from $Z$ is done independently, it follows that:
\begin{equation*}
\mathbb{P}(v_i \models_{Z,G}^{ps} v_j|G=g) \leq (1-p_2)^{B_c^g} p_2^{B_{nc}^g}.
\end{equation*}
Our goal, however, is not to bound  $\mathbb{P}(v_i \models_{Z,G} v_j|G=g)$, but rather $\mathbb{P}(v_i \models_{Z,G} v_j|(v_i,v_j) \notin E)$:
\begin{equation*}
\begin{split}
&\mathbb{P}(v_i \models_{Z,G} v_j|(v_i,v_j) \notin E)\\
\leq &\mathbb{P}(v_i \models_{Z,G}^{ps} v_j|(v_i,v_j) \notin E)\\
= & \sum_{g \in G^*_{i,j}}{\mathbb{P}(v_i \models_{Z,G}^{ps} v_j|G=g)\mathbb{P}(G=g|G \in G^*_{i,j})}\\
\leq & \sum_{g \in G^*_{i,j}}{(1-p_2)^{B_c^g} p_2^{B_{nc}^g}\mathbb{P}(G=g|G \in G^*_{i,j})}\\
= & \mathbb{E}[(1-p_2)^{B_c} p_2^{B_{nc}}|G \in G^*_{i,j}].
\end{split}
\end{equation*}
By Lemma \ref{lem:binom}, $B_c$ and $B_{nc}$ are independent of $(v_i,v_j) \notin E$, which is equivalent to $G \in G^*_{i,j}$; thus, our bound is equal to $\mathbb{E}[(1-p_2)^{B_{c}} p_2^{B_{nc}}]$. Also by Lemma \ref{lem:binom}, $B_c$ is a binomial random variable with parameters $|V|-j$ and $p_1^2$ while $B_{nc}$ is a binomial random variable with parameters $j-2$ and $p_1^2$; furthermore, $B_c$ and $B_{nc}$ are independent of each other. Therefore, our bound becomes  $\mathbb{E}[(1-p_2)^{B_{c}} p_2^{B_{nc}}]=\mathbb{E}[(1-p_2)^{B_{c}}]\mathbb{E}[p_2^{B_{nc}}]$. We can now use the moment generating function of the binomial to get:
\begin{equation*}
\begin{split}
\mathbb{E}[(1-p_2)^{B_{c}}]&=\mathbb{E}[e^{(\ln{(1-p_2)})B_{c}}]\\
&=(1-p_1^2+p_1^2e^{\ln{(1-p_2)}})^{|V|-j}\\
&=(1-p_1^2+(1-p_2)p_1^2)^{|V|-j},\\
\mathbb{E}[p_2^{B_{nc}}]&=\mathbb{E}[e^{(\ln{p_2})B_{c}}]\\
&=(1-p_1^2+p_1^2e^{\ln{p_2}})^{j-2}\\
&=(1-p_1^2+p_2p_1^2)^{j-2}.
\end{split}
\end{equation*}
Taking the product gives 
\begin{equation*}
\begin{split}
&\mathbb{E}[(1-p_2)^{B_{c}}]\mathbb{E}[p_2^{B_{nc}}]\\
=&(1-p_1^2+(1-p_2)p_1^2)^{|V|-j}(1-p_1^2+p_2p_1^2)^{j-2}.
\end{split}
\end{equation*}
 \end{proof}
\noindent Replacing $(1-p_2)$ and $p_2$ with $\max{\{p_2,(1-p_2)\}}$ in the r.h.s., we get the weaker but simpler bound:
\begin{restatable}{corollary}{corsimple}\label{cor:simple}
Let $Z$ be chosen randomly from $2^{V_{i,j}}$ as follows: for every $v \in V_{i,j}$, we include $v \in Z$ with i.i.d. probability $0<p_2<1$. Then, $\mathbb{P}(v_i \models_{Z,G} v_j|(v_i,v_j) \notin E)$ is upper bounded by $(1-(1-\max{\{p_2,(1-p_2)\}})p_1^2)^{|V|-2}$.
\end{restatable}
\noindent Since $0<p_2<1$, $0<1-\max{\{p_2,(1-p_2)\}}<1$, and since $0<p_1<1$, $0<1-(1-\max{\{p_2,(1-p_2)\}})p_1^2<1$. Thus, for fixed $p_1$ and $p_2$, the bound in Corollary \ref{cor:simple} decays exponentially fast to $0$ as $|V| \rightarrow \infty$, and therefore so does the bound in Theorem \ref{thm:randomz}---as it is tighter. Therefore, the probability that a random $Z$ d-separates $v_i$ and $v_j$ quickly becomes very low as $|V| \rightarrow \infty$. In fact, the decay to $0$ is guaranteed not just for fixed $p_1$, but rather as long as $p_1=\Omega(\sqrt{\frac{\log{|V|}}{|V|}})$ (although, of course, the decay will no longer necessarily be exponential).\footnote{If $p_1=\Omega(\sqrt{\frac{\log{|V|}}{|V|}})$, then for large enough $|V|$ we get that $(1-(1-\max{\{p_2,(1-p_2)\}})p_1^2)^{|V|-2} \leq (1-c\frac{\log{|V|}}{|V|})^{|V|-2}$ for some constant $c>0$, and by LHospital's rule this expression goes to $0$ as $|V| \rightarrow \infty$.} We can also get a slightly tighter bound on the unconditional probability of d-separation:
\begin{restatable}{corollary}{corunconditional}\label{cor:unconditional}
Let $Z$ be chosen randomly from $2^{V_{i,j}}$ as follows: for every $v \in V_{i,j}$, we include $v \in Z$ with i.i.d. probability $0<p_2<1$. Then, $\mathbb{P}(v_i \models_{Z,G} v_j)$ is upper bounded by $(1-p_1)(1-p_1^2+(1-p_2)p_1^2)^{|V|-j}(1-p_1^2+p_2p_1^2)^{j-2}$.
\end{restatable}

Our second bound considers all conditioning sets up to a certain size. $S$ in this proof includes all length $2$ \underline{noncollider} paths between $v_i$ and $v_j$.
\begin{theorem}\label{thm:boundedsize}
Let $Z^{0.5p_1^2(j-2)} \subseteq 2^{V_{i,j}}$ be the collection of all subsets of size up to at most $0.5p_1^2(j-2)$. Then, we can upper bound $\mathbb{P}(\exists Z \in Z^{0.5p_1^2(j-2)} \text{ s.t. } v_i \models_{Z,G} v_j|(v_i,v_j) \notin E)$ by $e^{-\frac{0.25p_1^2(j-2)}{2}}$.
\end{theorem}
\begin{proof}
To get pseudoseparation, it is necessary to block all paths from $Q_{nc}$ that exist in the graph, which means include all of their middle nodes in $Z$. By Lemma $\ref{lem:binom}$, the minimum set size needed to do so is at least a binomial random variable $B_{nc}$ with parameters $p_1^2$ and $j-2$. Hence every subset $Z \subseteq V_{i,j}$ s.t. $|Z| < B_{nc}$ is not d-separating. Using the Chernoff bound for a binomial random variable, we get the bound $\mathbb{P}(B_{nc} \leq 0.5p_1^2(j-2)) \leq e^{-\frac{0.25 p_1^2 (j-2)}{2}}$. However, in the event that $B_{nc} > 0.5p_1^2(j-2)$, no conditioning set of size at most $0.5p_1^2(j-2)$ is d-separating.
\end{proof}
\noindent For any fixed $0<p_1<1$, $0<\gamma<1$, for all $j \geq \gamma |V|$, $0.5p_1^2(j-2) \geq 0.5p_1^2(\gamma |V|-2) \rightarrow \infty$ (linearly fast) and $e^{-\frac{0.25p_1^2(j-2)}{2}} \leq  e^{-\frac{0.25p_1^2(\gamma |V|-2)}{2}} \rightarrow 0$ (exponentially fast) as $|V| \rightarrow \infty$. It follows that as $|V|$ grows, for an arbitrarily large fraction of node pairs, the minimum d-separating set size is likely to increase linearly with $|V|$. As with our first bound, we can slightly improve the bound when allowing for $(v_i,v_j) \in E$:
\begin{restatable}{corollary}{corboundedsizeunconditional}\label{cor:boundedsizeunconditional}
Let $Z^{0.5p_1^2(j-2)} \subseteq 2^{V_{i,j}}$ be the collection of all subsets of size up to at most $0.5p_1^2(j-2)$. Then, $\mathbb{P}(\exists Z \in Z^{0.5p_1^2(j-2)} \text{ s.t. } v_i \models_{Z,G} v_j)$ is upper bounded by $(1-p_1)e^{-\frac{0.25p_1^2(j-2)}{2}}$.
\end{restatable}

Our third bound chooses the conditioning set uniformly at random among all sets of a fixed size. We do not use it in Section \ref{sec:Algorithms}, but it contributes to our understanding of the search space for d-separation. In this proof, our choice of $S$ is actually done in reverse: we fix the conditioning set, and consider $S$ to be the set of all length $2$ paths which would cause the conditioning set to fail pseudoblocking.
\begin{theorem}\label{thm:fixedsize}
Let $\alpha \in \{0,1,2,\ldots,|V|-2\}$. Let $Z$ be chosen uniformly at random among all subsets of size $\alpha$ of $V_{i,j}$. Then, $\mathbb{P}(v_i \models_{Z,G} v_j|(v_i,v_j) \notin E)$ is upper bounded by $(1-p_1^2(2-p_1^2)\frac{\alpha}{|V|-2})^{|V|-j}(1-p_1^2)^{j-\alpha-2}$.

\end{theorem}
\begin{proof}
Let $2^{V_{i,j}}_\alpha$ be the set of all subsets of $V_{i,j}$ of size exactly $\alpha$. Let $z \in 2^{V_{i,j}}_\alpha$. Let $m_{nc}(z)=|V_{nc} - z|$ be the number of nodes from $V_{nc}$ outside of $z$, and $m_c(z)=|z \cap V_c|$ be the number of nodes from $V_c$ inside $z$. $m_{nc}(z)$ is the number of paths in $Q_{nc}$ which, \underline{if exist}, violate pseudoseparation; similarly, $m_c(z)$ is the number of paths in $Q_c$ that, if exist, violate pseudoseparation. Pseudoblocking holds iff all paths counted by $m(z)=m_c(z)+m_{nc}(z)$ do not exist, and by Lemma \ref{lem:independence} each of them fails to exist with probability $1-p_1^2$ independently of the others and of whether $(v_i,v_j) \in E$. Thus:
\begin{equation*}
\begin{split}
\mathbb{P}(v_i \models_{z,G}^{ps} v_j|(v_i,v_j) \notin E)&=(1-p_1^2)^{m(z)},
 \end{split}
 \end{equation*}
 Then, using the fact that $Z$ is independent from whether or not $(v_i,v_j) \in E$, we get:
 \begin{equation*}
\begin{split}
&\mathbb{P}(v_i \models_{Z,G}^{ps} v_j|(v_i,v_j) \notin E)\\
=&\sum_{z \in 2^{V_{i,j}}_\alpha}{\mathbb{P}(v_i \models_{Z,G}^{ps} v_j|(v_i,v_j) \notin E,Z=z)\mathbb{P}(Z=z)}\\
=&\sum_{z \in 2^{V_{i,j}}_\alpha}{(1-p_1^2)^{m(z)}\mathbb{P}(Z=z)}\\
=&\mathbb{E}[(1-p_1^2)^{m(Z)}]
 \end{split}
 \end{equation*}

Let $M_c=m_c(Z)$, $M_{nc}=m_{nc}(Z)$. $M_c$ determines $M_{nc}$, as exactly $\alpha-M_{c}$ slots in $Z$ are given to nodes from $V_{nc}$, so the remaining $|V_{nc}|-(\alpha-M_c)=j-2-(\alpha-M_c)=M_c+(j-\alpha-2)$ nodes from $V_{nc}$ end up outside of $Z$: $M_{nc}=M_c+(j-\alpha-2)$. Therefore, $m(Z)=M_c+M_{nc}=2M_c+(j-\alpha-2)$. Thus, we get:
 \begin{equation*}
\begin{split}
&\mathbb{P}(v_i \models_{Z,G}^{ps} v_j|(v_i,v_j) \notin E)\\
=&\mathbb{E}[(1-p_1^2)^{2M_c+(j-\alpha-2)}]\\
=&\mathbb{E}[((1-p_1^2)^2)^{M_c}]\mathbb{E}[(1-p_1^2)^{j-\alpha-2}]\\
=&\mathbb{E}[((1-p_1^2)^2)^{M_c}](1-p_1^2)^{j-\alpha-2}\\
=&\mathbb{E}[e^{\ln{((1-p_1^2)^2)}M_c}](1-p_1^2)^{j-\alpha-2}
 \end{split}
 \end{equation*}
\noindent Since $Z$ is chosen uniformly at random from $2^{V_{i,j}}_\alpha$, $M_c$ can be thought of as drawing $\alpha$ nodes from the population $V_{i,j}$ of nodes, where $V_c$ of the nodes are defined as ``success" states (as in, every node from $V_c$ that we draw increases the value of $M_c$ by one). This means that $M_c$ is a hypergeometric random variable with population size $N=|V|-2$, $K=|V_c|=|V|-j$ success states, and number of draws $d=\alpha$. Thus, $\mathbb{E}[e^{\ln{((1-p_1^2)^2)}M_c}]$ can be derived from the moment generating function of the hypergeometric distribution. It is known that the moment generating function of the hypergeometric distribution with parameters $N$, $K$ and $d$ is upper bounded by the moment generating function of the binomial random variable with parameters $\frac{d}{N}$ and $K$ \cite{hoeffding1963probability}. Applying this result, we get:
 \begin{equation*}
\begin{split}
&\mathbb{E}[e^{\ln{((1-p_1^2)^2)}M_c}](1-p_1^2)^{j-\alpha-2}\\
\leq & ((1-\frac{d}{N})+(1-p_1^2)^2\frac{d}{N})^K(1-p_1^2)^{j-\alpha-2}\\
= & ((1-\frac{\alpha}{|V|-2})+(1-p_1^2)^2\frac{\alpha}{|V|-2})^{|V|-j}(1-p_1^2)^{j-\alpha-2}\\
= & (1-p_1^2(2-p_1^2)\frac{\alpha}{|V|-2})^{|V|-j}(1-p_1^2)^{j-\alpha-2}\\
 \end{split}
 \end{equation*}
 Therefore we have established:
 \begin{equation*}
 \begin{split}
 &\mathbb{P}(v_i \models_{Z,G} v_j|(v_i,v_j) \notin E)\\
 \leq&\mathbb{P}(v_i \models_{Z,G}^{ps} v_j|(v_i,v_j) \notin E)\\
 \leq&(1-p_1^2(2-p_1^2)\frac{\alpha}{|V|-2})^{|V|-j}(1-p_1^2)^{j-\alpha-2}.
 \end{split}
 \end{equation*}
\end{proof}

\section{Analysis of Constraint-Based Methods}\label{sec:Algorithms}

In this section, we discuss the implications of the bounds from Section \ref{sec:Bounds} on the performance of constraint-based methods. We begin by proving that extremely sparse graphs are rare. After that, we define precision and introduce the PC and UniformSGS Algorithms. We then analyze the average-case performance of constraint-based methods on large and non extremely sparse graphs. We first show that constraint-based methods provide low precision when restricted to small conditioning sets (which they usually are in practice). We also show that even without that restriction, the PC algorithm must suffer from either low precision or an exponential number of oracle calls. We finish with an analysis of UniformSGS, showing it makes an expected exponential number of oracle calls even when $(v_i,v_j) \notin E$.

\subsection{Extreme Sparsity}\label{subsec:extreme}

In some areas, sparsity is considered to be an unrealistic assumption \cite{petersen2022causal,shirokikh2022networks}. In this subsection, we show a theoretical sense in which sparsity restricts us to a very small family of graphs. For simplicity of presentation, assume throughout Section \ref{sec:Algorithms} that $p_1$ is a weakly monotonic function of $|V|$. We show that the case where $\lim_{|V| \rightarrow \infty}{p_1}=0$ is rare in a well-defined sense. Let $G^{d}$ be the set of graphs in $G^*$ with density at most $d$. Denoting the density of $G$ as $d_1$, we show that the expected ratio $\mathbb{E}[\frac{|G^{d_1}|}{|G^*|}] \rightarrow 0$ as $|V| \rightarrow \infty$. We provide a deterministic lemma, and convert it into a probabilistic theorem.
\begin{restatable}{lemma}{lemsparse}\label{lem:sparse}
If $\lim_{|V| \rightarrow \infty}{d}=0$, then $\lim_{|V| \rightarrow \infty}{\frac{|G^{d}|}{|G^*|}}=0$.
\end{restatable}
\begin{restatable}{theorem}{thmsparse}\label{thm:sparse}
Let $d_1$ be the density of $G$. If $\lim_{|V| \rightarrow \infty}{p_1}=0$, then  $\lim_{|V| \rightarrow \infty}{\mathbb{E}[\frac{|G^{d_1}|}{|G^*|}]}=0$. Note that for any $\alpha>0$, this implies $\lim_{|V| \rightarrow \infty}{\mathbb{P}(\frac{|G^{d_1}|}{|G^*|}>\alpha)}=0$.
\end{restatable}

\subsection{Precision and the Algorithms Considered}\label{sec:algdef}

Before we discuss the performance of algorithms, we need to define precision. For any causal discovery algorithm, we denote the algorithm's prediction for $E$ as $E_{pred}$. Since the d-separation oracle is never wrong, $E \subseteq E_{pred}$ for any constraint-based method, as the algorithm would only remove an edge when it finds a d-separating set. Therefore, the only relevant type of mistake is predicting $(v_i,v_j) \in E_{pred}$ while $(v_i,v_j) \notin E$, which happens when the algorithm fails to find a d-separating set despite the fact that one exists.
\begin{definition}\label{def:precision}
Let $E_{pred} \in V \times V$ be the output of some causal discovery algorithm $A$ when the underlying causal graph is $G$. For any $v_i,v_j \in V$, we define the {\it precision} of $A$ on $v_i, v_j$ as $\mathbb{P}((v_i,v_j) \notin E_{pred}|(v_i,v_j) \notin E)$.
\end{definition}

Next, we define UniformSGS and PC \cite{spirtes1989causality,spirtes1991algorithm}: see Algorithms \ref{alg:uniformsgs} and \ref{alg:pc}. We focus on the skeleton (undirected graph) recovery portion of the algorithms, and ignore their edge orientation phase. Note that in Algorithm \ref{alg:pc}, we did not specify the order of the selected sets in the for loop.
\begin{algorithm}[th]
\caption{The UniformSGS Algorithm}
\label{alg:uniformsgs}
\begin{algorithmic}[1] 
\STATE $E_{pred} \gets \{(x,y):x,y \in V, x \neq y\}$
\FOR{$x,y \in V$}
	\FOR{$Z \subseteq V-\{x,y\}$ chosen uniformly at random without replacement}
		  \IF{$x \models_{Z,G} y$}
		  	\STATE{Remove $(x,y)$ from $E_{pred}$}
			\STATE{Break}
		  \ENDIF
	\ENDFOR
\ENDFOR
\RETURN $E_{pred}$
\end{algorithmic}
\end{algorithm}
\begin{algorithm}[th]
\caption{The PC Algorithm}
\label{alg:pc}
\textbf{Parameter}: $C_{\max} \in \mathbb{N} \cup \{\infty\}$
\begin{algorithmic}[1] 
\STATE $E_{pred} \gets \{(x,y):x,y \in V, x \neq y\}$
\STATE $\forall w \in V, Adj(w) \gets V-\{w\}$
\FOR{$x,y \in V$}
	\FOR{$Z \in 2^{Adj(x)-\{y\}} \cup 2^{Adj(y)-\{x\}}$ s.t. $|Z| \leq C_{\max}$}
		  \IF{$x \models_{Z,G} y$}
		  	\STATE{Remove $(x,y)$ from $E_{pred}$}
			\STATE{$Adj(x) \gets Adj(x)-\{y\}$}
    			\STATE{$Adj(y) \gets Adj(y)-\{x\}$}
			\STATE{Break}
		  \ENDIF
	\ENDFOR
\ENDFOR
\RETURN $E_{pred}$
\end{algorithmic}
\end{algorithm}

\subsection{Performance}\label{sec:perf}

For the remainder of this section, we assume that $\lim_{|V| \rightarrow \infty}{p_1} \neq 0$, which due to monotonicity means that $p_1$ is bounded from below by some positive constant. In practice, constraint based methods must replace the d-separation oracle with a statistical conditional independence test. Most of the tests used in practice are only accurate for small conditioning sets \cite{runge2019detecting,li2020nonparametric}, and thus most constraint-based methods restrict themselves to small conditioning sets in practice. However, a direct application of Theorem \ref{thm:boundedsize} shows that for a constraint-based method to achieve high precision, the size of the conditioning sets considered must increase linearly with $|V|$, which for large graphs leads to much larger conditioning sets than what's considered feasible by most statistical tests. Note that since $\delta_1$ in Corollary \ref{cor:arbitrary} below is arbitrary, the bound applies to any arbitrary fraction of all variable pairs.
\begin{restatable}{corollary}{corarbitrary}\label{cor:arbitrary}
Let $0<\delta_1<1$. Let $A$ be a constraint-based method which only calls the oracle with conditioning sets of size less than $0.5p_1^2(\delta_1|V|-2)$. Then, for every node pair $v_i, v_j \in V$ where $i<j$ and $j>\delta_1|V|$, we get that $\mathbb{P}((v_i,v_j) \notin E_{pred}|(v_i,v_j) \notin E)=O(e^{-|V|})$.
\end{restatable}

We now remove the statistical consideration and again assume we can call the oracle on conditioning sets of any size. For the PC Algorithm, Corollary \ref{cor:arbitrary} already shows that $C_{max}$ must grow linearly with $|V|$ for good precision. However, this doesn't rule out the possibility of a sweet spot: a value of $C_{\max}$ large enough to avoid Corollary \ref{cor:arbitrary} yet small enough to enable polynomial running time. Nevertheless, the upcoming Theorem \ref{thm:pc} and its Corollary \ref{cor:pcimplication} rule out such a sweet spot. Note that PC restricts the search space to $2^{Adj(x)-\{y\}} \cup 2^{Adj(y)-\{x\}}$  instead of $2^{V-\{x,y\}}$: this does not impact our analysis in part (i) of Theorem \ref{thm:pc}, but we do need to take it into consideration when we prove part (ii). We note that Theorem \ref{thm:pc} continues to hold even if PC searches in $2^{V-\{x,y\}}$ instead of $2^{Adj(x)-\{y\}} \cup 2^{Adj(y)-\{x\}}$.
\begin{restatable}{theorem}{thmpc}\label{thm:pc}
Let $0<\delta_1<1,0<\delta_2<1$. In the PC Algorithm:
\begin{enumerate}
\item[(i)] Assume $C_{\max} < 0.5p_1^2(\delta_1|V|-2)$. For every node pair $v_i, v_j \in V$ where $i<j$ and $j>\delta_1|V|$, we get $\mathbb{P}((v_i,v_j) \notin E_{pred}|(v_i,v_j) \notin E)=O(e^{-|V|})$.
\item[(ii)] Assume instead $C_{\max} > 0.5p_1(\delta_2|V|-2)$. Then, for every node pair $v_i,v_j \in V$, if $(v_i,v_j) \in E$ then with probability $1-O(e^{-|V|})$, the PC Algorithm makes $\Theta(e^{|V|})$ oracle calls for $x=v_i$, $y=v_j$. 
\end{enumerate}
\end{restatable}
We now formalize our claim that for large $|V|$, PC has low precision (case (i)) or exponential running time (case (ii)). In Corollary \ref{cor:pcimplication} below, note that s ince we can choose $\delta_1$ as small as we like to begin with, we can make sure that when we end up in case (i), it holds for an arbitrarily large fraction of the pairs. 
\begin{restatable}{corollary}{corpcimplication}\label{cor:pcimplication}
Let $0<\delta_1<1$. In the PC Algorithm, for every node pair $v_i, v_j \in V$ where $i<j$ and $j>\delta_1|V|$, either $\mathbb{P}((v_i,v_j) \notin E_{pred}|(v_i,v_j) \notin E)=O(e^{-|V|})$,  or the PC Algorithm makes $\Theta(e^{|V|})$ oracle calls with probability $1-O(e^{-|V|})$ for $x=v_i$, $y=v_j$ when $(v_i,v_j) \in E$.
\end{restatable}

Finally, we analyze UniformSGS. Since there is no size limit on the conditioning sets considered, then whenever $x=v_i$, $y=v_j$, $(v_i,v_j) \in E$, UniformSGS trivially requires an exponential number of oracle calls, because it is called on every subset in $2^{V_{i,j}}$. In Theorem \ref{thm:sgs} we show that for large $|V|$ we can usually expect an exponential number of oracle calls even when $(v_i,v_j) \notin E$.
\begin{restatable}{theorem}{thmsgs}\label{thm:sgs}
When testing whether $(v_i,v_j) \in E$, let $C$ be the number of oracle calls made by UniformSGS. Let $\alpha=1+\frac{1}{(2-p_1^2)^{|V|-2}}$.
\begin{enumerate}
\item[(i)]  $\mathbb{E}[C|(v_i,v_j) \notin E] \geq \frac{1}{\alpha} ((\frac{2}{2-p_1^2})^{|V|-2}-1)$.
\item[(ii)] $\mathbb{E}[C] \geq p_1 2^{|V|-2}+(1-p_1)\frac{1}{\alpha} ((\frac{2}{2-p_1^2})^{|V|-2}-1)$.
\end{enumerate}
\end{restatable}

\section{Conclusion}\label{sec:conclusion}

In this paper, we considered a random DAG model $G=(V,E)$, where each undirected edge is generated i.i.d. with a fixed probability. We have shown that on this model, even when d-separation is guaranteed to exist, only very few conditioning sets are d-separating. We have shown that as $|V| \rightarrow \infty$, unless the graph is extremely sparse, a randomly chosen conditioning set is highly unlikely to be d-separating, under different randomization scenarios. Specifically, when the conditioning set includes each node with any fixed i.i.d. probability, or when it is limited to fixed size, or even when we get to try all conditioning sets of size up to a certain (linear) fraction of the nodes---in all those cases, the probability of d-separation decays exponentially to $0$ as the $|V| \rightarrow \infty$. We showed that extremely sparse graphs represent a vanishingly small portion of all possible graphs. We used our bounds to show that in the average case, the PC and UniformSGS Algorithm, as well as any constraint-based method restricted to small conditioning sets, are likely to have poor performance (either in terms of precision or time) on large and non extremely sparse graphs. Our results highlight the need for a sophisticated search for d-separation in any constraint-based method meant for causal discovery in large graphs.
\bibliographystyle{abbrv}
\bibliography{dsepbibliography.org.tug}

\begin{thebibliography}{10}

\bibitem{chickering2002optimal}
D.~M. Chickering.
\newblock Optimal structure identification with greedy search.
\newblock {\em Journal of machine learning research}, 3(Nov):507--554, 2002.

\bibitem{ghassami2018multi}
A.~Ghassami, N.~Kiyavash, B.~Huang, and K.~Zhang.
\newblock Multi-domain causal structure learning in linear systems.
\newblock {\em Advances in neural information processing systems}, 31, 2018.

\bibitem{glymour2019review}
C.~Glymour, K.~Zhang, and P.~Spirtes.
\newblock Review of causal discovery methods based on graphical models.
\newblock {\em Frontiers in genetics}, 10:524, 2019.

\bibitem{heckerman2006bayesian}
D.~Heckerman, C.~Meek, and G.~Cooper.
\newblock A bayesian approach to causal discovery.
\newblock {\em Innovations in Machine Learning: Theory and Applications}, pages
  1--28, 2006.

\bibitem{heinze2018causal}
C.~Heinze-Deml, M.~H. Maathuis, and N.~Meinshausen.
\newblock Causal structure learning.
\newblock {\em Annual Review of Statistics and Its Application}, 5:371--391,
  2018.

\bibitem{hoeffding1963probability}
W.~Hoeffding.
\newblock Probability inequalities for sums of bounded random variables.
\newblock {\em Journal of the American statistical association},
  58(301):13--30, 1963.

\bibitem{hoyer2008nonlinear}
P.~Hoyer, D.~Janzing, J.~M. Mooij, J.~Peters, and B.~Sch{\"o}lkopf.
\newblock Nonlinear causal discovery with additive noise models.
\newblock {\em Advances in neural information processing systems}, 21, 2008.

\bibitem{huang2022latent}
B.~Huang, C.~J.~H. Low, F.~Xie, C.~Glymour, and K.~Zhang.
\newblock Latent hierarchical causal structure discovery with rank constraints.
\newblock {\em arXiv preprint arXiv:2210.01798}, 2022.

\bibitem{huang2018generalized}
B.~Huang, K.~Zhang, Y.~Lin, B.~Sch{\"o}lkopf, and C.~Glymour.
\newblock Generalized score functions for causal discovery.
\newblock In {\em Proceedings of the 24th ACM SIGKDD international conference
  on knowledge discovery \& data mining}, pages 1551--1560, 2018.

\bibitem{huang2020causal}
B.~Huang, K.~Zhang, J.~Zhang, J.~Ramsey, R.~Sanchez-Romero, C.~Glymour, and
  B.~Sch{\"o}lkopf.
\newblock Causal discovery from heterogeneous/nonstationary data.
\newblock {\em The Journal of Machine Learning Research}, 21(1):3482--3534,
  2020.

\bibitem{li2020nonparametric}
C.~Li and X.~Fan.
\newblock On nonparametric conditional independence tests for continuous
  variables.
\newblock {\em Wiley Interdisciplinary Reviews: Computational Statistics},
  12(3):e1489, 2020.

\bibitem{pearl2009causality}
J.~Pearl.
\newblock {\em Causality}.
\newblock Cambridge university press, 2009.

\bibitem{petersen2022causal}
A.~H. Petersen, J.~Ramsey, C.~T. Ekstr{\o}m, and P.~Spirtes.
\newblock Causal discovery for observational sciences using supervised machine
  learning.
\newblock {\em arXiv preprint arXiv:2202.12813}, 2022.

\bibitem{runge2019detecting}
J.~Runge, P.~Nowack, M.~Kretschmer, S.~Flaxman, and D.~Sejdinovic.
\newblock Detecting causal associations in large nonlinear time series
  datasets, sci. adv., 5, eaau4996, 2019.

\bibitem{scholkopf2021toward}
B.~Sch{\"o}lkopf, F.~Locatello, S.~Bauer, N.~R. Ke, N.~Kalchbrenner, A.~Goyal,
  and Y.~Bengio.
\newblock Toward causal representation learning.
\newblock {\em Proceedings of the IEEE}, 109(5):612--634, 2021.

\bibitem{shimizu2006linear}
S.~Shimizu, P.~O. Hoyer, A.~Hyv{\"a}rinen, A.~Kerminen, and M.~Jordan.
\newblock A linear non-gaussian acyclic model for causal discovery.
\newblock {\em Journal of Machine Learning Research}, 7(10), 2006.

\bibitem{shirokikh2022networks}
O.~Shirokikh, G.~Pastukhov, A.~Semenov, S.~Butenko, A.~Veremyev, E.~L.
  Pasiliao, and V.~Boginski.
\newblock Networks of causal relationships in the us stock market.
\newblock {\em Dependence Modeling}, 10(1):177--190, 2022.

\bibitem{spirtes1991algorithm}
P.~Spirtes and C.~Glymour.
\newblock An algorithm for fast recovery of sparse causal graphs.
\newblock {\em Social science computer review}, 9(1):62--72, 1991.

\bibitem{spirtes1989causality}
P.~Spirtes, C.~Glymour, and R.~Scheines.
\newblock Causality from probability.
\newblock 1989.

\bibitem{SGS93}
P.~Spirtes, C.~Glymour, and R.~Scheines.
\newblock {\em Causation, Prediction, and Search}.
\newblock Spring-Verlag Lectures in Statistics, 1993.

\bibitem{spirtes2016causal}
P.~Spirtes and K.~Zhang.
\newblock Causal discovery and inference: concepts and recent methodological
  advances.
\newblock In {\em Applied informatics}, volume~3, pages 1--28. SpringerOpen,
  2016.

\bibitem{spirtes2013causal}
P.~L. Spirtes, C.~Meek, and T.~S. Richardson.
\newblock Causal inference in the presence of latent variables and selection
  bias.
\newblock {\em arXiv preprint arXiv:1302.4983}, 2013.

\bibitem{verma1988influence}
T.~Verma and J.~Pearl.
\newblock {\em Influence diagrams and d-separation}.
\newblock UCLA, Computer Science Department, 1988.

\bibitem{vowels2022d}
M.~J. Vowels, N.~C. Camgoz, and R.~Bowden.
\newblock D’ya like dags? a survey on structure learning and causal
  discovery.
\newblock {\em ACM Computing Surveys}, 55(4):1--36, 2022.

\bibitem{xie2020generalized}
F.~Xie, R.~Cai, B.~Huang, C.~Glymour, Z.~Hao, and K.~Zhang.
\newblock Generalized independent noise condition for estimating latent
  variable causal graphs.
\newblock {\em Advances in neural information processing systems},
  33:14891--14902, 2020.

\bibitem{yao2022temporally}
W.~Yao, G.~Chen, and K.~Zhang.
\newblock Temporally disentangled representation learning.
\newblock In A.~H. Oh, A.~Agarwal, D.~Belgrave, and K.~Cho, editors, {\em
  Advances in Neural Information Processing Systems}, 2022.

\bibitem{zhang2012identifiability}
K.~Zhang and A.~Hyvarinen.
\newblock On the identifiability of the post-nonlinear causal model.
\newblock {\em arXiv preprint arXiv:1205.2599}, 2012.

\end{thebibliography}

\newpage
\begin{appendices}
\section{Proofs}\label{app:proofs}

In this appendix, we include the proofs omitted from the main paper.

\corsimple*
\begin{proof}
Immediate from Theorem \ref{thm:randomz}, since by replacing $p_2$ and $(1-p_2)$ with their upper bound $\max{\{p_2,(1-p_2)\}}$, we get:
\begin{equation*}
\begin{split}
& (1-p_1^2+(1-p_2)p_1^2)^{|V|-j}(1-p_1^2+p_2p_1^2)^{j-2}\\
\leq & (1-p_1^2+\max{\{p_2,(1-p_2)\}}p_1^2)^{|V|-j+j-2}\\
= & (1-p_1^2+\max{\{p_2,(1-p_2)\}}p_1^2)^{|V|-2}\\
= & (1-(1-\max{\{p_2,(1-p_2)\}})p_1^2)^{|V|-2}
\end{split}
\end{equation*}
\end{proof}

\corunconditional*
\begin{proof}
Recall from Theorem \ref{thm:dsepequiv} that $v_i$ and $v_j$ are d-separable in $G$ iff $(v_i,v_j) \notin E$, so $\mathbb{P}(v_i \models_{Z,G} v_j|(v_i,v_j) \in E)=0$. Therefore, we get:
\begin{equation*}
\begin{split}
&\mathbb{P}(v_i \models_{Z,G} v_j)\\
=&\mathbb{P}(v_i \models_{Z,G} v_j|(v_i,v_j) \notin E)(1-p_1)+0 \cdot p_1\\
=&\mathbb{P}(v_i \models_{Z,G} v_j|(v_i,v_j) \notin E)(1-p_1).
\end{split}
\end{equation*}
Applying the bound from Theorem \ref{thm:randomz} completes the proof.
\end{proof}

\corboundedsizeunconditional*
\begin{proof}
Similar to Corollary \ref{cor:unconditional}.
\end{proof}

\lemsparse*

\begin{proof}
Assume $\lim_{|V| \rightarrow \infty}{d}=0$. Consider $|V|$ large enough so that $d<0.5$, and for convenience assume $\frac{1}{d}$ divides $\binom{|V|}{2}$. The total number of graphs in $G^*$ is
\begin{equation*}
A=2^{\binom{|V|}{2}}.
\end{equation*}
For any $n$, the number of graphs with exactly $n$ edges is $\binom{\binom{|V|}{2}}{n}$. Graphs in $G^*$ with density at most $d$ have at most $d\binom{|V|}{2}$ edges; the number of such graphs is therefore
\begin{equation*}
B=\sum_{n=0}^{d\binom{|V|}{2}}{\binom{\binom{|V|}{2}}{n}}.
\end{equation*}
We claim that $B \leq \frac{d}{1-2d-\binom{|V|}{2}^{-1}} A$ (it's actually true that $B \leq dA$, but we don't need it); if we can show this, then the ratio of graphs with density at most $d$ to all graphs is
\begin{equation*}
\frac{B}{A} \leq \frac{\frac{d}{1-2d-\binom{|V|}{2}^{-1}}A}{A}=\frac{d}{1-2d-\binom{|V|}{2}^{-1}},
\end{equation*}
which goes to $0$ as $|V| \rightarrow \infty$, so that will prove our claim.

Why does $B \leq \frac{d}{1-2d-\binom{|V|}{2}^{-1}}A$? Note that by definition, $A=\sum_{n=0}^{\binom{|V|}{2}}{\binom{\binom{|V|}{2}}{n}}$. Therefore, $B$ is the first $d\binom{|V|}{2}$ terms of $A$. Let $C=\sum_{n=d\binom{|V|}{2}+1}^{(1-d)\binom{|V|}{2}-1}{\binom{\binom{|V|}{2}}{n}}$. Every term in $C$ is larger than every term in $B$ (binomial coefficients are larger the closer they are to the midpoint), and the number of terms in $C$ is $(1-2d)\binom{|V|}{2}-1$, while the number of terms in $B$ is $d\binom{|V|}{2}$. It follows that $\frac{B}{A} < \frac{B}{C} < \frac{d}{1-2d-\binom{|V|}{2}^{-1}}$.
\end{proof}

\thmsparse*

\begin{proof}
Assume $\lim_{|V| \rightarrow \infty}{p_1}=0$. Condition on $d_1 < \sqrt{p_1}$ to get
\begin{equation*}
\begin{split}
&\mathbb{E}[\frac{|G^{d_1}|}{|G^*|}]\\
=&\mathbb{E}[\frac{|G^{d_1}|}{|G^*|}|d_1 < \sqrt{p_1}]\mathbb{P}(d_1 < \sqrt{p_1})\\
&+\mathbb{E}[\frac{|G^{d_1}|}{|G^*|}|d_1 \geq \sqrt{p_1}]\mathbb{P}(d_1 \geq \sqrt{p_1}).
\end{split}
\end{equation*}
We bound the first term using $\mathbb{P}(d_1 < \sqrt{p_1}) \leq 1$:
\begin{equation*}
\begin{split}
&\mathbb{E}[\frac{|G^{d_1}|}{|G^*|}|d_1 < \sqrt{p_1}]\mathbb{P}(d_1 < \sqrt{p_1})\\
\leq & \mathbb{E}[\frac{|G^{d_1}|}{|G^*|}|d_1 < \sqrt{p_1}]\\
\leq & \frac{|G^{\sqrt{p_1}}|}{|G^*|}.
\end{split}
\end{equation*}
Then, we bound the second term via the fact that $\frac{|G^{d_1}|}{|G^*|} \leq 1$ for all possible values of $d_1$:
\begin{equation*}
\begin{split}
&\mathbb{E}[\frac{|G^{d_1}|}{|G^*|}|d_1\ \geq\sqrt{p_1}] \mathbb{P}(d_1 \geq \sqrt{p_1}) \\ 
\leq&\mathbb{P}(d_1 \geq \sqrt{p_1}).
\end{split}
\end{equation*}
Using Markov's inequality and the fact that $\mathbb{E}[d_1]=p_1$, we get that
\begin{equation*}
\mathbb{P}(d_1 \geq \sqrt{p_1}) \leq \frac{\mathbb{E}[d_1]}{\sqrt{p_1}} = \sqrt{p_1}.
\end{equation*}
Combining the bounds, we get $\mathbb{E}[\frac{|G^{d_1}|}{|G^*|}] \leq \frac{|G^{\sqrt{p_1}}|}{|G^*|} + \sqrt{p_1}$. As $p_1 \rightarrow 0$, $\frac{|G^{\sqrt{p_1}}|}{|G^*|} \rightarrow 0$ by Lemma \ref{lem:sparse}. Since $\lim_{|V| \rightarrow \infty}{p_1}=0$, this completes the proof.
\end{proof}

\corarbitrary*

\begin{proof}
Follows from Theorem \ref{thm:boundedsize} and the fact that for $(v_i,v_j) \notin E_{pred}$ it is necessary that there exists a d-separating subset in $\{Z \in 2^{V_{i,j}}: |Z| \leq C_{\max}\}$, where $C_{\max}$ is the maximum size of conditioning set considered by the algorithm.
\end{proof}

\thmpc*

\begin{proof}
Part (i) follows from Theorem \ref{thm:boundedsize} and the fact that for $(v_i,v_j) \notin E_{pred}$ it is necessary that there exists a d-separating subset in $\{Z \in 2^{V_{i,j}}: |Z| \leq C_{\max}\}$. For part (ii), consider $(v_i,v_j) \in E$. Edges from $E$ never get deleted, so at all times $E \subseteq E_{pred}$. Defining $S_j=Adj(v_j)-\{v_i\}$, $|S_j|$ is therefore always bounded from below by the number of edges in $E$ adjacent to $v_j$ except $(v_i,v_j)$, which is a binomial random variable with parameters $p_1$ and $|V|-2$. Using the Chernoff bound, we get that 
\begin{equation*}
\mathbb{P}(|S_j| \leq 0.5p_1(|V|-2)) \leq e^{-\frac{0.25p_1(|V|-2)}{2}}=O(e^{-|V|}).
\end{equation*}
Therefore, with probability $1-O(e^{-|V|})$,
\begin{equation*}
|S_j|>0.5p_1(|V|-2)>0.5p_1(\delta_2|V|-2),
\end{equation*}
and since also $C_{\max}>0.5p_1(\delta_2 |V|-2)$, there are at least $2^{0.5p_1(\delta_2|V|-2)}=\Theta(e^{|V|})$ subsets in the search space (we can use $\Theta$ instead of $\Omega$ because there are at most $2^{|V|-2}=O(e^{|V|})$ subsets). Since $(v_i,v_j) \in E$, there is no d-separating set, so the oracle will be called on every subset in the search space.
\end{proof}

\corpcimplication*

\begin{proof}
Note that no matter how small we choose $\delta_1>0$ in Theorem \ref{thm:pc}, we can always choose $\delta_2>0$ s.t. $(p_1\delta_1-\delta_2)|V|>2p_1-2$, or equivalently $p_1(\delta_1|V|-2)>\delta_2|V|-2$. When we do so, $0.5p_1^2(\delta_1|V|-2)>0.5p_1(\delta_2|V|-2)$, and so we are necessarily in either case (i) or (ii) of the theorem. This is the formalization of our claim that for large $|V|$, PC has low precision (case (i)) or exponential running time (case (ii)).
\end{proof}

\thmsgs*

\begin{proof}
First, assume $(v_i,v_j) \notin E$. Let $K=|\{z \in 2^{V-\{v_i,v_j\}}:v_i \cancel{\models}_{z,G} v_j\}|$ (the number of subsets that are not d-separating in $G$). Let $G^*_{i,j}=\{g=(V,E_g) \in G^*: (v_i,v_j) \notin E_g\}$, and for each $g \in G^*_{i,j}$, let $k_g$ be the value of $K$ when $G=g$. Conditional on $K=k_g$, $C$ is a negative hypergeometric random variable with population size $N=2^{|V|-2}$ (all subsets), $k_g$ success states, and failure number $r=1$. The expectation of such a negative hypergeometric random variable is $\frac{k_g}{N-k_g+1}$. Therefore, we get
\begin{equation*}
\begin{split}
&\mathbb{E}[C|(v_i,v_j) \notin E]\\
=&\mathbb{E}[C|G \in G^*_{i,j}]\\
=&\sum_{g \in G^*_{i,j}}{\mathbb{E}[C|G=g]\mathbb{P}(G=g|G \in G^*_{i,j})}\\
=&\sum_{g \in G^*_{i,j}}{\frac{k_g}{N-k_g+1}\mathbb{P}(G=g|G \in G^*_{i,j})}\\
=&\mathbb{E}[\frac{K}{N-K+1}|G \in G^*_{i,j}]\\
\geq &\frac{\mathbb{E}[K|G \in G^*_{i,j}]}{N-\mathbb{E}[K|G \in G^*_{i,j}]+1}\\
= &\frac{\mathbb{E}[K|(v_i,v_j) \notin E]}{N-\mathbb{E}[K|(v_i,v_j) \notin E]+1},
\end{split}
\end{equation*}
\noindent where the inequality follows from Jensen's inequality.

Suppose $Z$ is chosen as in Theorem \ref{thm:randomz} with $p_2=0.5$, meaning uniformly at random over $2^{V-\{v_i,v_j\}}$. Because $Z$ is chosen uniformly, $\mathbb{P}(v_i \cancel{\models}_{Z,G} v_j|G=g)$ is simply $\frac{k_g}{2^{|V|-2}}$, and so $k_g=2^{|V|-2}\mathbb{P}(v_i \cancel{\models}_{Z,G} v_j|G=g)$. Now we can compute:
\begin{equation*}
\begin{split}
&\mathbb{E}[K|(v_i,v_j) \notin E]\\
=&\mathbb{E}[K|G \in G^*_{i,j}]\\
=&\sum_{g \in G_{i,j}^*}{k_g\mathbb{P}(G=g|G \in G_{i,j}^*})\\
=&\sum_{g \in G_{i,j}^*}{2^{|V|-2}\mathbb{P}(v_i \cancel{\models}_{Z,G} v_j|G=g)\mathbb{P}(G=g|G \in G_{i,j}^*})\\
=&2^{|V|-2}\sum_{g \in G_{i,j}^*}{\mathbb{P}(v_i \cancel{\models}_{Z,G} v_j|G=g)\mathbb{P}(G=g|G \in G_{i,j}^*})\\
=&2^{|V|-2}\mathbb{P}(v_i \cancel{\models}_{Z,G} v_j| G \in G^*_{i,j})\\
=&2^{|V|-2}\mathbb{P}(v_i \cancel{\models}_{Z,G} v_j|(v_i,v_j) \notin E).
\end{split}
\end{equation*}
Applying Theorem \ref{thm:randomz}, we get $\mathbb{P}(v_i \cancel{\models}_{Z,G} v_j|(v_i,v_j) \notin E) \geq 1-(1-0.5p_1^2)^{|V|-2}$, and therefore we get:
\begin{equation*}
\begin{split}
&\mathbb{E}[K|(v_i,v_j) \notin E] \geq 2^{|V|-2}(1-(1-0.5p_1^2)^{|V|-2})\\
=&2^{|V|-2}-(2-p_1^2)^{|V|-2},
\end{split}
\end{equation*}
and thus, bringing everything together:
\begin{equation*}
\begin{split}
&\mathbb{E}[C|(v_i,v_j) \notin E]\\
\geq & \frac{\mathbb{E}[K|(v_i,v_j) \notin E]}{N-\mathbb{E}[K|(v_i,v_j) \notin E]+1}\\
=& \frac{\mathbb{E}[K|(v_i,v_j) \notin E]}{2^{|V|-2}-\mathbb{E}[K|(v_i,v_j) \notin E]+1}\\
\geq & \frac{2^{|V|-2}-(2-p_1^2)^{|V|-2}}{(2-p_1^2)^{|V|-2}+1}\\
=&\frac{2^{|V|-2}-(2-p_1^2)^{|V|-2}}{(2-p_1^2)^{|V|-2}+\frac{(2-p_1^2)^{|V|-2}}{(2-p_1^2)^{|V|-2}}}\\
=&\frac{2^{|V|-2}-(2-p_1^2)^{|V|-2}}{(1+\frac{1}{(2-p_1^2)^{|V|-2}})(2-p_1^2)^{|V|-2}}\\
=&\frac{2^{|V|-2}-(2-p_1^2)^{|V|-2}}{\alpha(2-p_1^2)^{|V|-2}}\\
 =&\frac{1}{\alpha} ((\frac{2}{2-p_1^2})^{|V|-2}-1).
\end{split}
\end{equation*}
We get the unconditional bound in (ii) with the following calculation:
\begin{equation*}
\begin{split}
&E[C]\\
=&E[C|(v_i,v_j) \notin E]\mathbb{P}((v_i,v_j) \notin E)\\
&+E[C|(v_i,v_j) \in E]\mathbb{P}((v_i,v_j) \in E)\\
=&E[C|(v_i,v_j) \notin E](1-p_1)+2^{|V|-2}p_1\\
\end{split}
\end{equation*}
\end{proof}

\section{Experiments}\label{app:experiments}

\begin{figure}[h]
\includegraphics[width=\linewidth,clip]{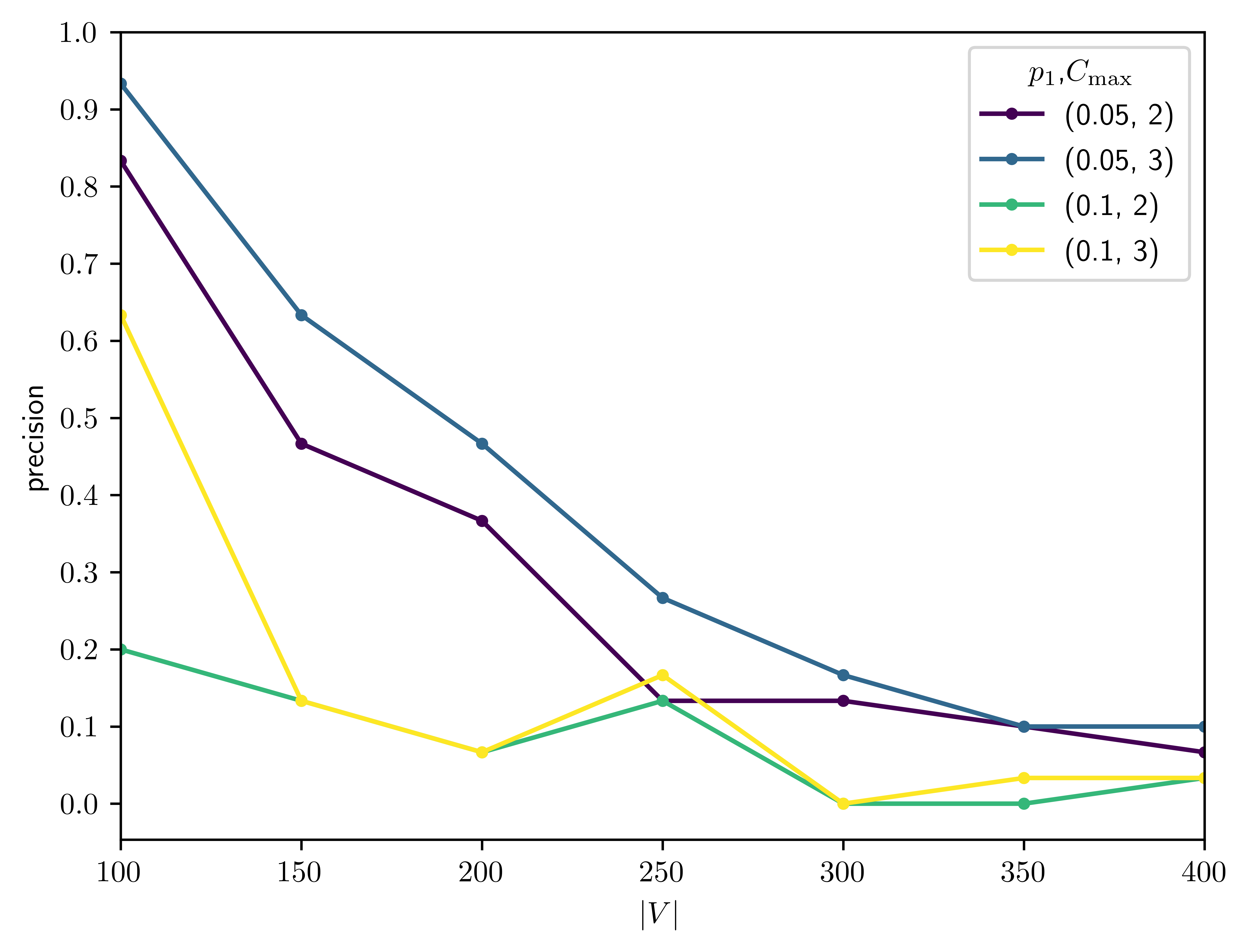}
\caption{Perfect PC experiment.}
\label{fig:experiment1}
\end{figure}

In this appendix, we present the results of two experiments. Our first experiment (Figure \ref{fig:experiment1}) tests the performance of the PC Algorithm, and more generally relates to Theorems \ref{thm:boundedsize} and \ref{thm:pc} as well as Corollaries \ref{cor:arbitrary} and \ref{cor:pcimplication}. In our experiment, given a random DAG with $|V|$ nodes and expected density $p_1 \in \{0.05,0.1\}$, we sample uniformly at random $30$ d-separable pairs without replacement, and test each pair for d-separation using {\bf all} conditioning sets of size up to $C_{\max} \in \{2,3\}$. We define the empirical precision to be the percentage of pairs (among the $30$) for which a d-separating conditioning set is found. To isolate the analysis of PC from any issues involving conditional independence test quality, data sample size, or any particular functional form of the structural equation model (SEM), we endow PC with an always-correct d-separation oracle. Even with this idealized perfect version of PC, our results show that PC performs poorly on our test graphs. Note that because we are using a perfect oracle, PC finds a d-separating set iff one with size at most $C_{\max}$ exists, and therefore this experiment also tests Theorem \ref{thm:boundedsize} and Corollary \ref{cor:arbitrary}; specifically, our experiment upper bounds the performance of any constraint-based method restricting itself to conditioning sets up to size at most $C_{\max}$. Consistently with Theorems \ref{thm:boundedsize} and \ref{thm:pc} as well as Corollaries \ref{cor:arbitrary} and \ref{cor:pcimplication}, precision quickly becomes very low.

\begin{figure}
\includegraphics[width=\linewidth,clip]{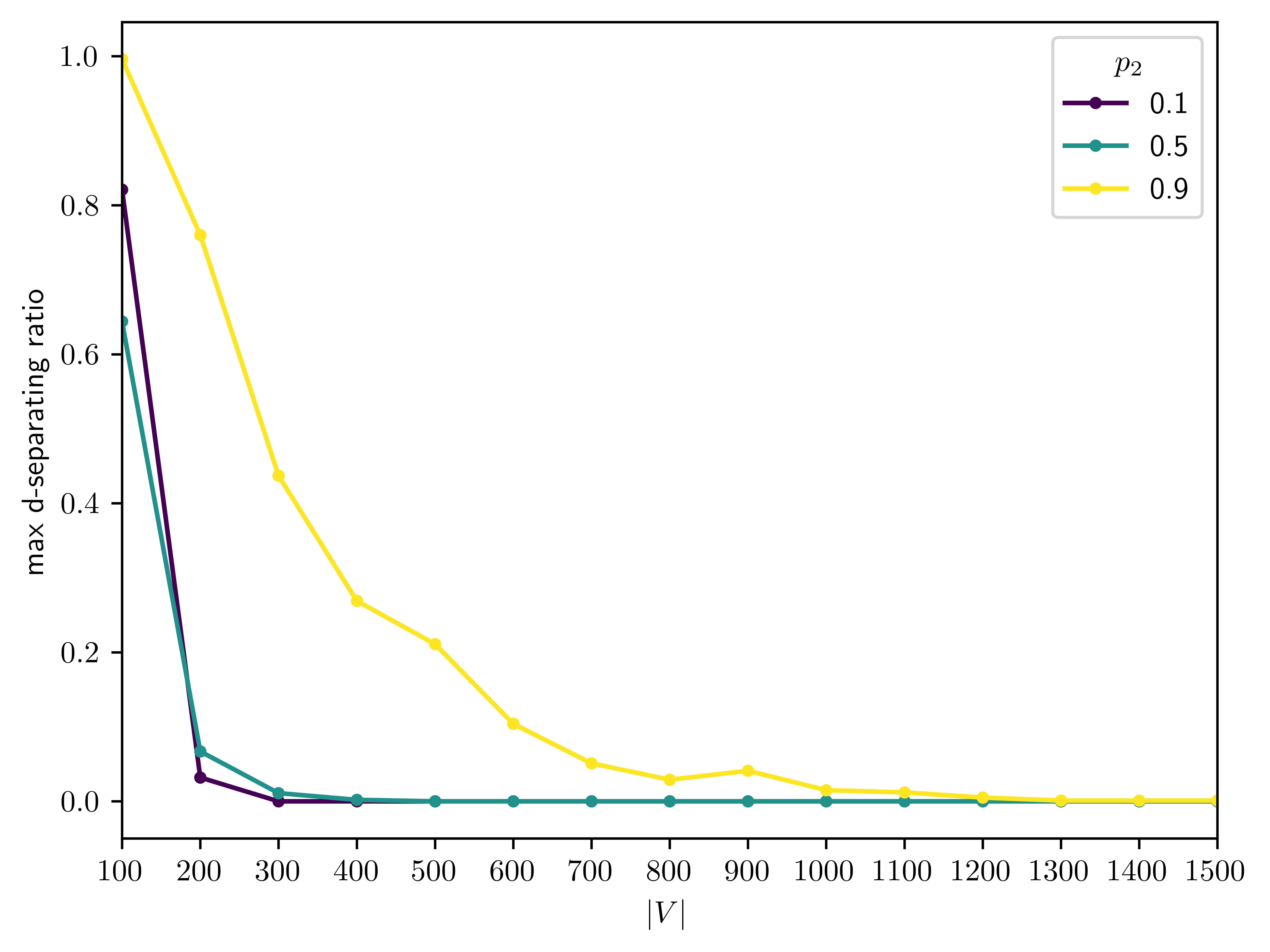}
\includegraphics[width=\linewidth,clip]{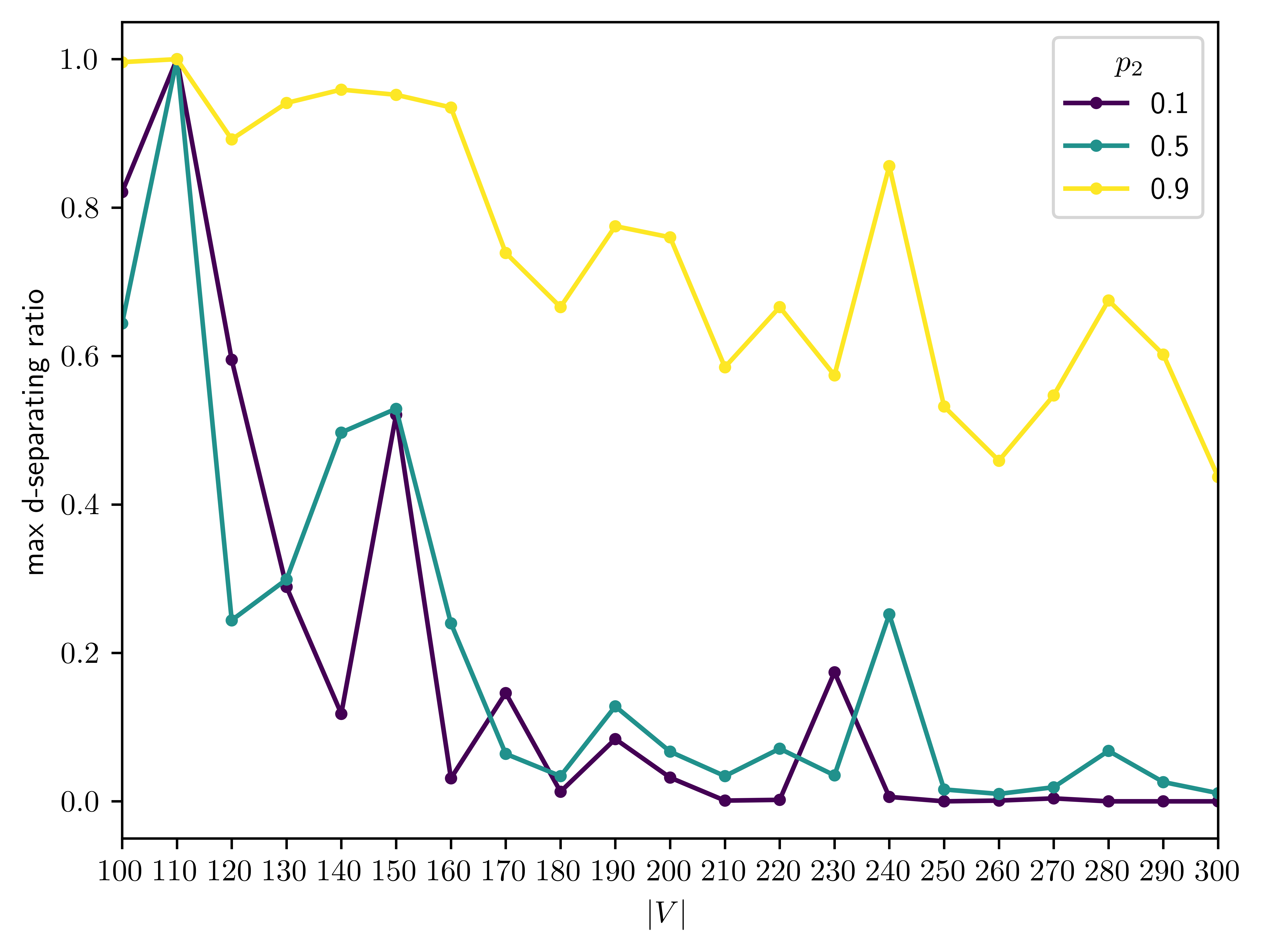}
\caption{d-separation probability ($p_1=0.05$). The bottom plot describes the same scenario as the top, but more fine-grained on the range $100 \leq |V| \leq 300$.}
\label{fig:experiment205}
\end{figure}

\begin{figure}
\includegraphics[width=\linewidth,clip]{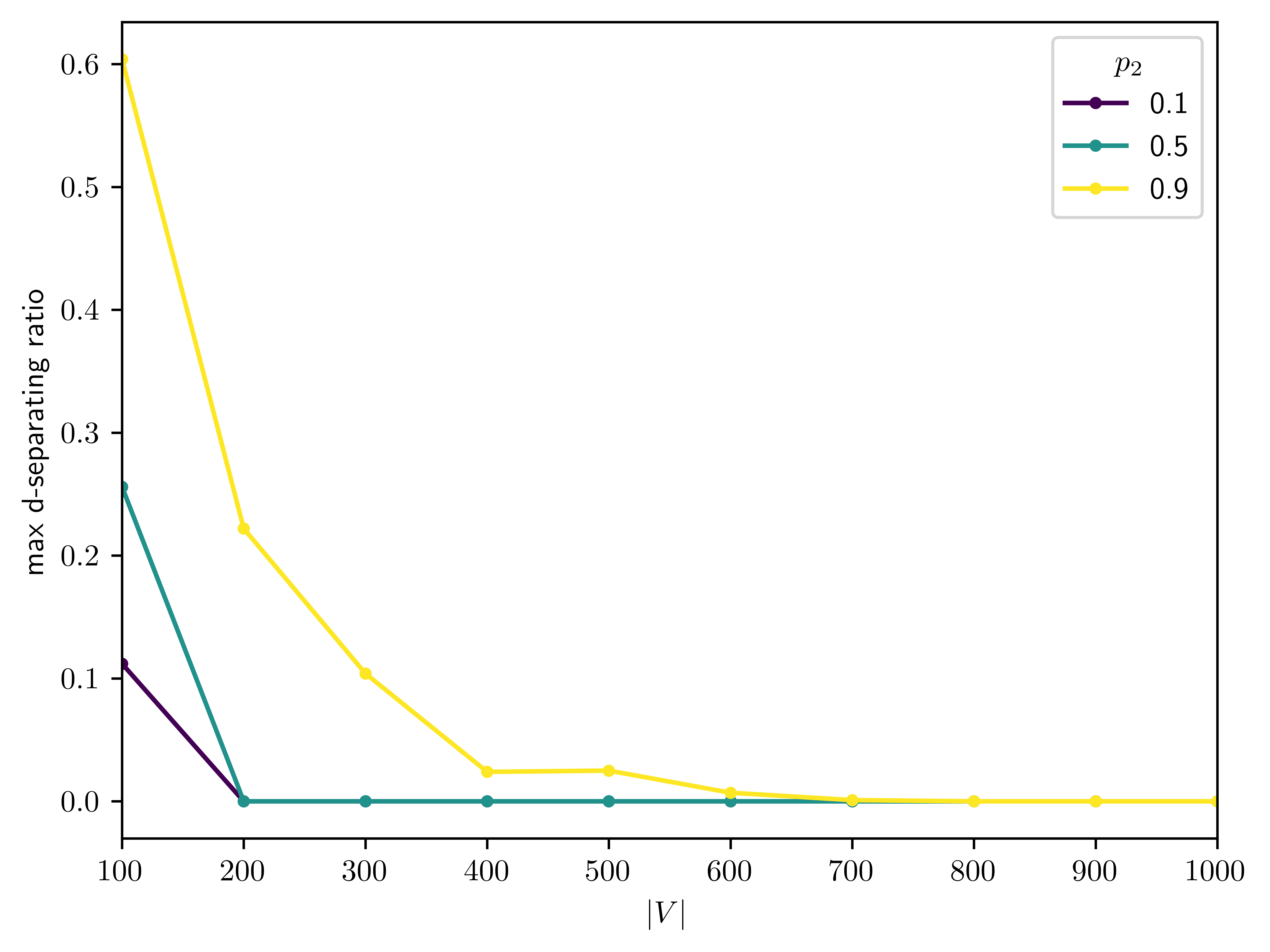}
\includegraphics[width=\linewidth,clip]{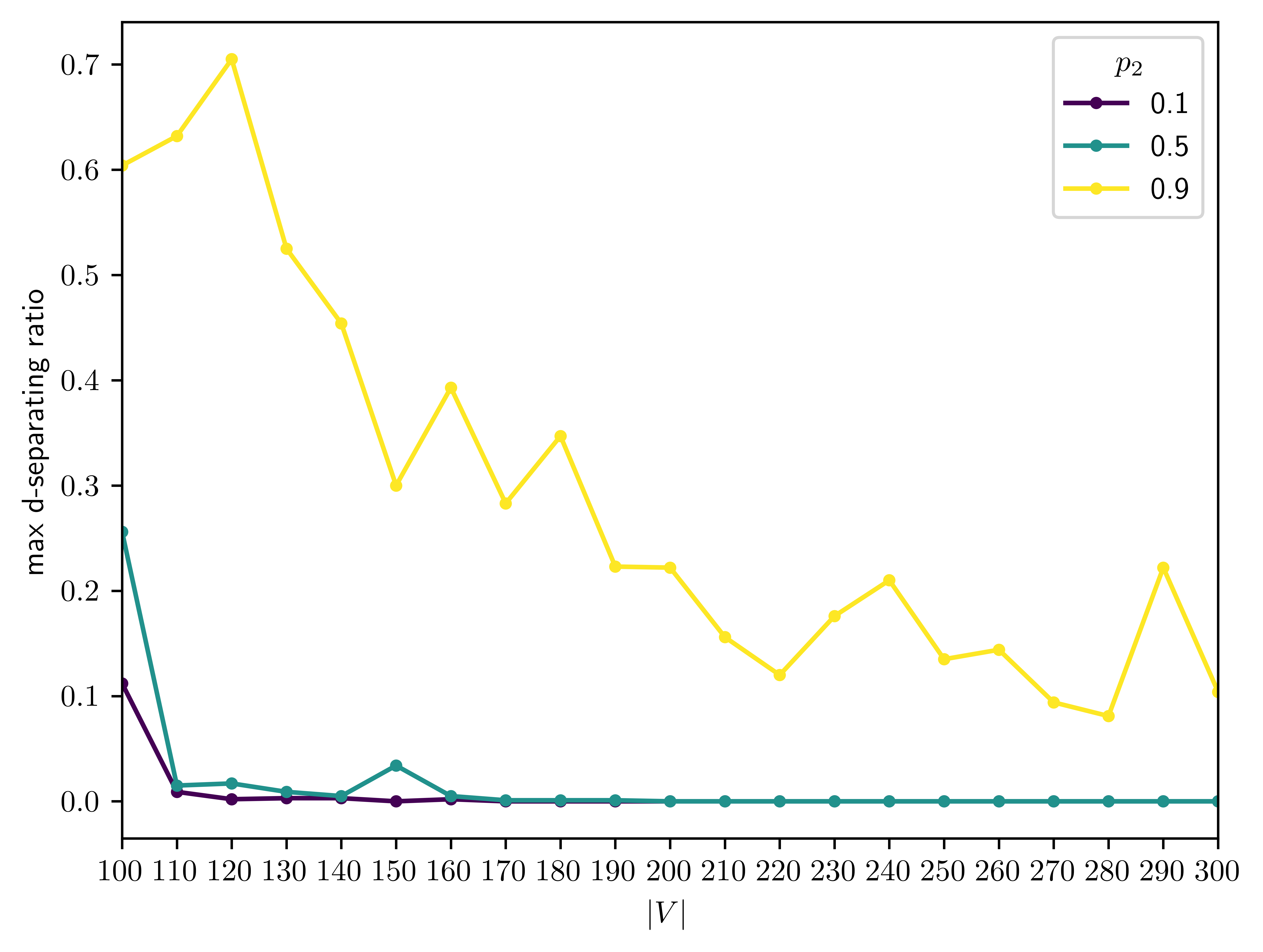}
\caption{d-separation probability ($p_1=0.1$). The bottom plot describes the same scenario as the top, but more fine-grained on the range $100 \leq |V| \leq 300$.}
\label{fig:experiment21}
\end{figure}

Our second experiment (Figures \ref{fig:experiment205} and \ref{fig:experiment21}, see next page) tests the bound in Theorem \ref{thm:randomz}. Given a random DAG with $|V|$ nodes and expected density $p_1 \in \{0.05, 0.1\}$, we sample uniformly at random $100$ d-separable variable pairs without replacement. For each pair, we sample $1000$ conditioning sets, including each variable (except the pair) in the conditioning set with i.i.d. probability $p_2$, and then calculate the ratio of the number of d-separating sets over all $1000$ sets. We then calculate the \underline{maximum} ratio over all $100$ pairs. Consistently with Theorem \ref{thm:randomz}, d-separation quickly becomes rare as $|V| \rightarrow \infty$ despite the low density.
\end{appendices}
\end{document}